%% file: main.tex
\documentclass{article}
\usepackage[T1]{fontenc}
\usepackage{microtype,graphicx,booktabs,float,hyperref,capt-of}

\usepackage[accepted]{icml2025}
\usepackage{amsmath,amssymb,amsthm,tabularx}
\makeatletter
\renewcommand{\Notice@String}{}
\renewcommand{\printAffiliationsAndNotice}[1]{%
  {\let\thefootnote\relax\footnotetext{\hspace*{-\footnotesep}%
  Correspondence: \href{mailto:miroslav.lzicar@deepmedchem.com}{miroslav.lzicar@deepmedchem.com}.\par
  #1}}
}
\makeatother
\hypersetup{
  pdfsubject={},
  pdfauthor={Miroslav Lžičař},
  pdftitle={OPDiv: Optimal Selection of Top-K High-Scoring, Diverse Compounds}
}

\newtheoremstyle{compactquestion}{10pt}{10pt}{\itshape}{}{\bfseries}{.}{0.5em}{}
\theoremstyle{compactquestion}
\newtheorem*{mainquestion}{Main Question}
\theoremstyle{plain}

\icmltitlerunning{OPDiv: Optimal Selection of High-Scoring, Diverse Compounds}
\begin{document}
\twocolumn[
\icmltitle{OPDiv: Optimal Selection of Top-K High-Scoring, Diverse Compounds}
\begin{icmlauthorlist}
\end{icmlauthorlist}
\begin{center}
{\bfseries Miroslav L\v{z}i\v{c}a\v{r}}\\
Deep MedChem
\end{center}
\icmlkeywords{Molecular diversity, portfolio optimization, compound selection, cheminformatics}
\vskip 0.3in
]
\printAffiliationsAndNotice{Preprint draft, September 2026.}

\begin{abstract}
A virtual screening campaign may produce thousands of promising candidates, but only a small number can be purchased, synthesized, or tested. The practical question is how to select a set of compounds that both rank well and are diverse enough: this poses a genuine tradeoff, where selecting the highest-scoring molecules yields limited diversity, while diversity selection sacrifices some well-scoring molecules. We introduce OPDiv, a diversity selection and evaluation algorithm solving this tradeoff by finding an optimal subset of molecules using integer optimization. We demonstrate the selection algorithm in practice with fingerprint distance, shape and electrostatic diversity and compare the resulting diversity spectra. We argue that virtual screening is not merely a ranking problem, but also an implicit constrained optimization task: when redundant chemotypes are undesirable, pipelines should be compared based on the top-k compound selections satisfying the desired diversity constraints. OPDiv makes it possible to find the optimal compound set under a given diversity threshold efficiently and serves as a fair benchmark of the best diverse selection achievable by a given structure-based or ligand-based virtual screening pipeline, molecular search or generative model. 
\end{abstract}

\begin{center}
\footnotesize
\begin{tabular}{@{}ll@{}}
\toprule
\textbf{Resource} & \textbf{Link or command} \\
\midrule
GitHub & \href{https://github.com/mireklzicar/opdiv}{github.com/mireklzicar/opdiv} \\
PyPI & \href{https://pypi.org/project/opdiv/}{pypi.org/project/opdiv} \\
Install & \texttt{pip install opdiv} \\
\bottomrule
\end{tabular}
\end{center}

\section{Introduction}
 To discover a potent chemical series, compound selection should balance promising analogues with alternative chemotypes or scaffold hops. This is related to the exploration--exploitation trade-off: prioritizing currently promising candidates versus exploring alternative regions of chemical space \citep{langevin2024}. Recent methodological advances make this problem all the more pressing for practitioners: ultra-large chemical spaces as well as DNA-encoded libraries may face problems with limited diversity or combinatorial explosion \citep{garciaortegon2025,martin2020}; machine-learning methods along with generative AI often exploit locally promising clusters and may get stuck in local optima or reinforcement-learning induced modal collapse \citep{blaschke2020memory,renz2024}. 

\noindent\begin{minipage}{\linewidth}
\raggedright
\begin{mainquestion}
Can we select compounds for highest scores and diversity in an optimal way? How can we evaluate such selections?
\end{mainquestion}
\centering
\includegraphics[width=0.9\linewidth]{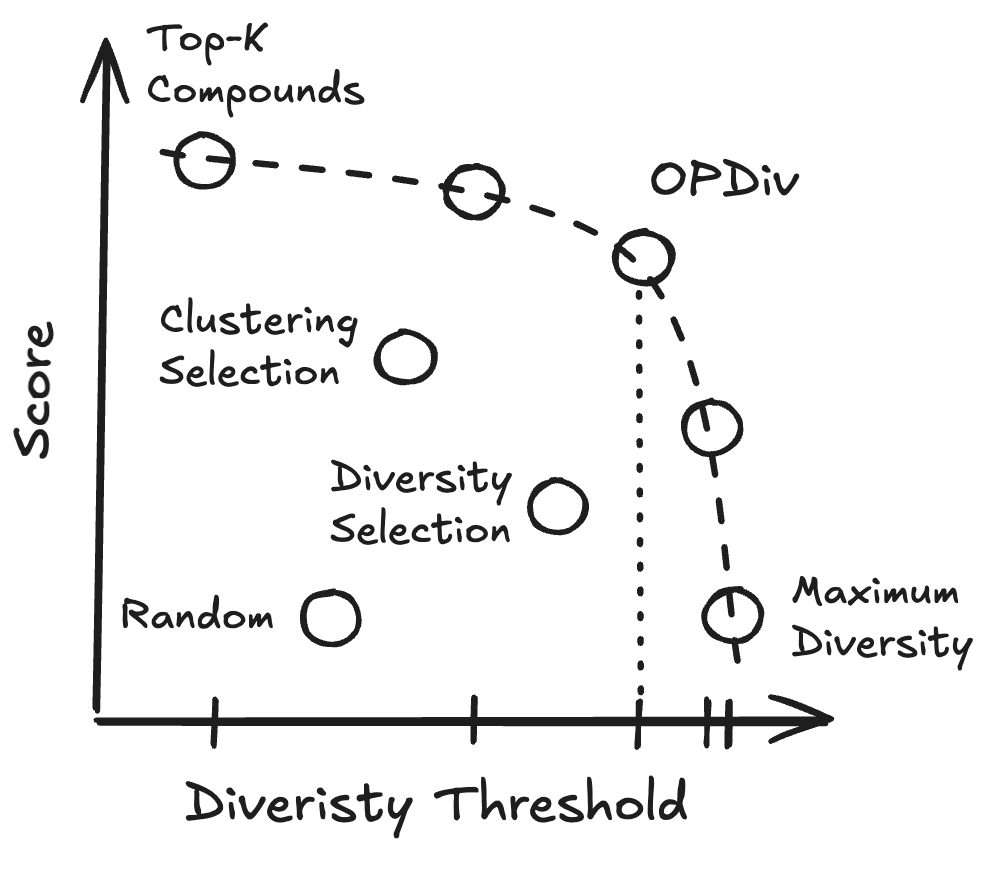}
\captionof{figure}{Schematic score–diversity trade-off. The goal of OPDiv is to find a subset of highest-scoring compounds from virtual screening at a given diversity threshold (such as Tanimoto similarity, shape, electrostatic or scaffold distance). A low minimum-distance threshold permits ordinary top-$k$ selection, whereas increasing the threshold requires greater separation and may eventually make a size-$k$ selection infeasible. This can also be seen as a variation of a multi-property optimization task, where OPDiv ultimately covers the achievable Pareto frontier.}
\label{fig:opdiv-cartoon}
\end{minipage}

\begin{table*}[t]
\centering
\small
\setlength{\tabcolsep}{5pt}
\renewcommand{\arraystretch}{1.08}

\caption{
Overview of diversity measures, subset-selection methods, and
score--diversity criteria considered in this work.
}
\label{tab:comparison}

\begin{tabularx}{\textwidth}{
@{}
>{\raggedright\arraybackslash}p{0.17\textwidth}
>{\raggedright\arraybackslash}p{0.17\textwidth}
>{\raggedright\arraybackslash}p{0.10\textwidth}
>{\raggedright\arraybackslash}X
@{}}

\toprule
\textbf{Metric / method} &
\textbf{Reference} &
\textbf{Type} &
\textbf{Summary} \\
\midrule

IntDiv &
\citet{benhenda2017} &
Diversity &
Mean pairwise molecular dissimilarity \\

SEDiv &
\citet{thomas2021} &
Diversity &
Fraction retained by sphere exclusion \\

\#Clusters &
\citet{butina1999} &
Diversity &
Number of clusters at a chosen similarity resolution \\

\#Circles &
\citet{xie2023} &
Diversity &
Maximum size of a threshold-separated subset \\

DISE &
\citet{gobbi2015} &
Selection &
Directed sphere exclusion; candidates prioritized by score \\

MaxMin &
\citet{ashton2002} &
Selection &
Greedy maximization of the minimum distance to the selected set \\

MaxSum &
\citet{borodin2012} &
Selection &
Greedy maximization of summed distance to the selected set \\

OptiSim &
\citet{clark1997} &
Selection &
Sampled MaxMin selection combined with sphere exclusion \\

DivCount\textsuperscript{\dag} &
\citet{renz2024} &
Criterion &
Number of threshold-separated hits above a score cutoff \\

GPDiv &
This work &
Criterion &
Mean score obtained by greedy diverse selection \\

CPDiv &
This work &
Criterion &
Best achievable mean score under cluster capacities \\

OPDiv &
This work &
Criterion &
Best achievable mean score of a diverse selection \\

\bottomrule
\end{tabularx}

\par\smallskip
\parbox{\textwidth}{\footnotesize\raggedright
\textsuperscript{\dag}Shorthand introduced here.
}

\end{table*}

A typical in silico screening pipeline consists of \textbf{1.} ranking and filtering (think of docking scores, molecular dynamics, machine learning predictions, medchem descriptors, QSAR or ADMET); and \textbf{2.} some sort of diversity selection: clustering of molecules, deduplicating scaffolds or sphere-exclusion heuristics. This can also be a trivial \textit{top-k} best selection or a random choice depending on more or less arbitrary factors such as price or availability. We imagine the position of these approaches in the score-diversity tradeoff as drawn in Figure \ref{fig:opdiv-cartoon}. Our contribution is making this selection optimal. Furthermore, we are then able to fairly evaluate in silico scoring pipelines (such as molecular search, active learning or de novo generative methods) based on how well they can explore chemical space in terms of finding the best scores, while maintaining diversity.

OPDiv thus connects these two tasks. For compound selection, it identifies the highest-scoring set that meets the chosen size and diversity requirements. For evaluation, the mean score of that set measures what the candidate pool makes available for experimental follow-up. The optimization uses established methods; the central contribution is the common criterion for selecting compounds and evaluating their source pool. Varying the diversity requirement shows the trade-off between score and separation.

\section{Selecting a high-scoring, diverse set}
\label{sec:background}
\subsection{Diversity Selection Background}

Portfolio optimization provides a useful perspective on compound selection: the aim is to choose a collection whose members offer both individual value and complementary opportunities. In classical financial portfolio theory, expected return is balanced against risk, with portfolio variance depending on covariances between asset returns \citep{markowitz1952}. Related formulations have been applied to molecular selection \citep{yevseyeva2019}, and mean--variance reasoning has been used to motivate diversity in molecular design \citep{langevin2024}. For compound selection, the motivation is that closely related molecules may share liabilities, making several promising analogs less informative or less protective against failure than their individual scores suggest.

OPDiv adopts a discrete portfolio-selection formulation: it maximizes the mean score of a fixed-size subset under explicit diversity constraints. Molecular similarity controls redundancy; it is not treated as a calibrated estimate of outcome covariance. The resulting score--diversity curve describes the best attainable score at each diversity requirement. Its interpretation concerns the supplied scores and molecular representations, rather than a direct estimate of experimental success or financial-style risk.

Existing selection methods include: dissimilarity selection (reviewed by \citet{gillet2011}); threshold-separated reagent selection and clique detection by \citet{gardiner1998}; maximum-score diversity selection which combines predicted activity with diversity \citep{meinl2010,meinl2011} and portfolio optimization \citep{yevseyeva2019}. \citet{qin2012} develop greedy and exact graph algorithms for diversified top-$k$ retrieval. PrefDiv combines relevance and diversity \citep{ge2020}, while SPARROW additionally considers synthesis \citep{fromer2025}. Internal diversity summarizes average dissimilarity; cluster counts and \#Circles measure grouping or packing; sphere-exclusion diversity reports a retained fraction; and diverse-hit counts measure separated candidates above a score cutoff \citep{benhenda2017,thomas2021,xie2023,renz2024}. Our research follows in this line of work. Table~\ref{tab:comparison} summarizes the metrics; definitions and baseline algorithms are collected in Appendix~\ref{app:baselines}.

Chemotype-aware evaluation also has an established history in virtual screening. \citet{mackey2009} analyze cluster-average metrics, which weight each active inversely by the number of actives in its chemotype, and first-found metrics, which credit only the earliest retrieved active from each chemotype. They show that first-found metrics can depend strongly on chemotype sizes and can lose sensitivity to differences in screening performance. This distinction is relevant here: measuring retrieval across active chemotypes and measuring the quality of the best available experimental shortlist address different evaluation objectives.

\label{sec:methods}
\subsection{Optimal and greedy selection}
\label{sec:opdiv}
Start with eligible, distinct molecules and a score for each. Higher scores indicate more desirable candidates; a lower-is-better score, such as a docking score, can be negated. Choose the number of compounds to advance, $N$, and a molecular similarity measure. Two compounds may be selected together only when their similarity is at most $\tau$. The score measures desirability; the similarity measure controls redundancy.

We call a set of $N$ distinct compounds selected jointly for experimental follow-up a \emph{molecular portfolio}. In the default formulation, each compound occupies one selection slot and contributes equally to the portfolio's mean score; diversity requirements constrain which compounds may be selected together.

\textbf{Optimal Portfolio Diversity (OPDiv) is the highest mean score achievable by selecting exactly $N$ molecules that satisfy the diversity requirement.} The selected portfolio's mean score evaluates the candidate pool. With $q_i$ denoting molecule $i$'s score,
\begin{equation}
 Q_N(\tau)=\max_{\substack{\text{allowed sets }S\\|S|=N}}\frac1N\sum_{i\in S}q_i.
 \label{eq:quality}
\end{equation}
An allowed set contains only eligible, distinct molecules, and every selected pair has similarity at most $\tau$. If no allowed set of size $N$ exists, the request is infeasible and no size-$N$ value is reported. A minimum-score filter can exclude weak candidates before selection.

\phantomsection\label{sec:gpdiv}
\textbf{Greedy Portfolio Diversity (GPDiv) is the mean score obtained by a score-ordered greedy selection under the same requirements.} Scan molecules from highest to lowest score, accepting each one that is sufficiently different from every molecule already selected. Stop after $N$ acceptances or after exhausting the pool; ties follow a fixed order. This is an established sphere-exclusion procedure \citep{qin2012,gobbi2015}. When it returns $N$ compounds, GPDiv cannot exceed OPDiv. Otherwise, no size-$N$ GPDiv value is assigned.

The best individual molecule need not belong to the best set. Consider four compounds, A--D, with illustrative docking scores of $-16$, $-12$, $-11$, and $-6$, where more negative is better. Only A--B and A--C are too similar to select together. To choose two compounds, greedy selection takes A first and then D, giving a mean docking score of $-11$. Selecting B and C instead gives the better mean of $-11.5$. Without D, greedy selection stops at A even though B and C still form an allowed pair.

\subsection{Allowing several analogs per series}
\label{sec:clustered}
Some selections allow several analogs while limiting the number from each chemical series. \textbf{Clustered Portfolio Diversity (CPDiv) is the highest mean score achievable for $N$ molecules under an upper limit on the number selected from each fixed cluster.} Every molecule belongs to exactly one cluster. These cluster limits replace the pairwise similarity rule. Score-based selection within chemical groups has established precedents \citep{oashi2011,jansen2019}.

For these disjoint groups and upper limits alone, greedy selection is exact: scan molecules by decreasing score and accept each while its cluster has space. Equivalently, keep the best members of each cluster up to its limit, then take the overall top $N$. Every such subset respects the limits, and any lower-scoring member of a cluster can be replaced by a retained higher-scoring member. This is the classical greedy solution for a partition matroid \citep{rado1957,edmonds1971}. If fewer than $N$ molecules remain, the request is infeasible.

A limit of one chooses the best-scoring representative from each selected group. Larger limits allow several analogs per series. Groups may also be defined by scaffolds or generic frameworks. Report the group definition, limits, and selected counts. Compounds in different clusters can still be similar, and reclustering a growing pool can change the selection rule. Overlapping groups, minimum quotas, or additional pairwise exclusions require a different optimization method.

\subsection{Fixed chemotype weights}
\label{sec:weighted-opdiv}
The formulation also accommodates predefined chemotype priorities. Inverse cluster-size weighting has been used to reduce the influence of abundant active series in virtual-screening evaluation \citep{clark2008,mackey2009}. Let $g(i)$ identify molecule $i$'s chemotype and let $\omega_j\geq0$ be a fixed weight for group $j$. Making the candidate pool $A$ explicit, the weighted objective is
\begin{equation}
 Q_N^{(\omega)}(A;\tau)
 =\max_{\substack{S\subseteq A,\ |S|=N\\S\text{ allowed at }\tau}}
 \frac1N\sum_{i\in S}\omega_{g(i)}q_i.
 \label{eq:weighted-opdiv}
\end{equation}
Replacing $q_i$ with $\omega_{g(i)}q_i$ preserves the linear objective and the existing selection constraints. Under CPDiv's disjoint groups and upper capacities alone, greedy selection remains exact when ordered by weighted score.

For example, $\omega_j=1/c_j$, where $c_j>0$ is a group's size in a fixed reference collection, gives less weight to members of abundant groups. Structural group sizes and counts of known actives define different weightings. With retrospective activity labels, inverse active-chemotype counts recover the contributions used in cluster-average enrichment, up to normalization (Appendix~\ref{app:weighted-enrichment}). Withheld activity labels must remain outside prospective selection. Group assignments and weights must be fixed across compared pools; recomputing them as a pool grows changes the objective and can invalidate its monotonicity. The experiments below use equal weights.

\section{Evaluating the candidate pool}
\label{sec:archive-evaluation}
\subsection{Credit for the best available shortlist}
To evaluate a molecular search for a follow-up task, apply the same final selection criterion to every candidate pool. This measures which method made the strongest diverse shortlist available. The acquisition budget measures resources spent finding and scoring candidates; $N$ measures how many compounds can advance. Comparisons require consistent acquisition budgets, eligibility rules, score scales, molecular representations, and diversity requirements.

A pool can contain a strong shortlist even when most of its molecules are repetitive. Adding candidates cannot make the best available set worse, provided existing scores and pairwise relationships stay fixed: the previous selection remains available. The exact optimum is also unchanged by input order or duplicate records removed before selection. Unselected or duplicate outputs can still consume acquisition resources and must be accounted for consistently. These properties concern the exact criterion; a time-limited solver may return different approximations.

The best-shortlist interpretation should be distinguished from a claim about general screening accuracy. As emphasized by \citet{mackey2009}, evaluation based on the best-ranked member of each chemotype can be sensitive to chemotype abundance. OPDiv deliberately evaluates the strongest feasible subset of a given pool, and additional distinct analogs can improve that subset by providing more opportunities for a high score. It therefore does not remove the effects of pool composition or score error. Comparisons of molecular search methods require matched acquisition budgets, while claims about experimental hit or chemotype recovery require independent activity labels.

Two score differences answer different practical questions. The ordinary top-$N$ mean minus OPDiv is the \emph{cost of requiring diversity}. OPDiv minus the mean of another allowed size-$N$ selection is the \emph{score lost by that selection procedure}. The first is unavoidable under the chosen requirement; the second could be recovered from the same pool without weakening it. When optimization leaves a gap between the bounds, these differences are reported as ranges.

\subsection{The score--diversity trade-off}
\label{sec:frontier}
A single similarity limit gives one answer to the selection problem. Repeating the calculation over a common range of limits gives a \emph{score--diversity curve}: the best achievable mean score at each diversity requirement. For Tanimoto similarity, a minimum distance of $\delta$ corresponds to a maximum similarity of $1-\delta$. Stronger separation can only lower the optimum or make the requested set infeasible, because it removes allowed choices.

The curve shows how much score must be given up for greater separation and whether a comparison depends on one particular cutoff. If two candidate pools' curves cross, their relative value depends on the diversity required. Summarizing the curve by an area or weighted average requires choosing a range and weights.

A different view asks how scores are distributed \emph{within} one selected set. Sort its scores from highest to lowest to obtain a \emph{selected-set score profile}, also called a ranked diversity spectrum. Each point represents one compound. The profile reveals whether a high mean is sustained across the set or conceals weak members. Each requested size is optimized separately. The first ten members of an optimal 50-compound set need not form the optimal ten-compound set, and equal optimal means can arise from different profiles.

\subsection{An enrichment factor based on random acquisition}
\label{sec:random-equivalent}
Conventional enrichment factors compare active recovery with its expectation under random selection, while early-recognition metrics emphasize retrieval near the top of a ranking \citep{truchon2007,mackey2009}. We propose a complementary question for OPDiv: how many randomly acquired compounds would be needed to supply an equally strong diverse portfolio?

Let $A$ be a candidate pool obtained with acquisition budget $B$, and let $q_\star=Q_N(A;\tau)$ be its feasible OPDiv value. Choose a reference library $\mathcal U$ of $M$ eligible, distinct compounds with fixed scores and pairwise relationships. In this finite-library definition, $B$ counts distinct eligible compounds acquired and scored. For a uniformly random permutation of $\mathcal U$, let $R_m$ contain its first $m$ compounds. Define $T(q_\star)$ as the smallest $m$ for which $R_m$ contains an allowed size-$N$ portfolio with mean score at least $q_\star$. Both the candidate and random pools receive the same final selection criterion.

The \emph{OPDiv random-equivalent sampling factor} is
\begin{equation}
 \mathrm{EF}_{\mathrm{OPDiv}}(A;\mathcal U,B)
 =\frac{\mathbb E[T(q_\star)]}{B},
 \label{eq:opdiv-enrichment}
\end{equation}
where the expectation is over random acquisition orders. A value of eight means that random acquisition requires eight times the candidate method's budget on average to match its portfolio. A reliability-based alternative reports
\begin{equation}
 m_p(q_\star)=\min\{m:\Pr[T(q_\star)\leq m]\geq p\},
 \label{eq:matching-budget-quantile}
\end{equation}
for example at $p=0.5$ and $0.9$.

This factor compares sampling effort rather than dividing score values. Its interpretation depends on the reference library, acquisition policy, portfolio size, and diversity requirement. An unreachable target is reported as such. Other acquisition-cost conventions require matched random baselines. Appendix~\ref{app:random-equivalent} describes estimation, solver uncertainty, and analytic checks. This sampling factor is a proposed extension and is not evaluated in the experiments below.

\section{Implementation and optimization guarantees}
\label{sec:implementation}
\label{sec:certificates}
The implementation takes molecular records and finite scores, removes duplicates and ineligible candidates, and identifies pairs that are too similar to select together. Identity handling must specify salts, tautomers, stereochemistry, and preparation states so that repeated records of one experimental compound cannot occupy several places. Pairwise rules can use fingerprints, learned embeddings, or explicit three-dimensional comparisons. No triangle inequality is required, but asymmetric comparisons need symmetrization, and explicit 3D comparisons need conformer, protonation, and alignment policies.

OPDiv uses Python, RDKit representations, and the OR-Tools CP-SAT solver. Binary variables indicate which compounds are selected; one constraint fixes the number of compounds and another excludes each forbidden pair. This is an established integer optimization problem \citep{qin2012}; the formulation is given in Appendix~\ref{app:computation}. The output includes the selected molecular identities, their original scores, and an interval $L\leq Q_N\leq U$. Here $L$ is the best validated selection mean and $U$ is an upper bound on what any allowed selection could achieve. Their difference states how much improvement may remain. At the time limit, the solver may leave an optimality gap or feasibility unresolved.

Computation can focus on the highest-scoring part of a pool while still bounding the full pool. Optimistic placeholders account for omitted candidates, and the bounds include integer-score rounding (Appendix~\ref{app:computation}). Selected identities, counts, original scores, and every selected pair are checked independently. We retain the selected compounds and solver bounds for each result.

\section{Retrospective evaluation}
\label{sec:selector-results}
\subsection{Data and comparison protocol}
\label{sec:selector-protocol}
We use molecules generated by Augmented Memory in the diverse-hit benchmark of \citet{renz2024} for DRD2, GSK3, and JNK3, with four independent runs per target. We take the first 10,000 unique acquisitions from each run, keep valid structures with finite scores, and remove duplicates by canonical identity. Morgan fingerprints use radius 2, 2,048 bits, and no chirality. We compute them from the original molecular graphs and use canonical SMILES to identify duplicates. Scores retain their original higher-is-better scale.

\begin{figure*}[t]
\centering
\includegraphics[width=\textwidth]{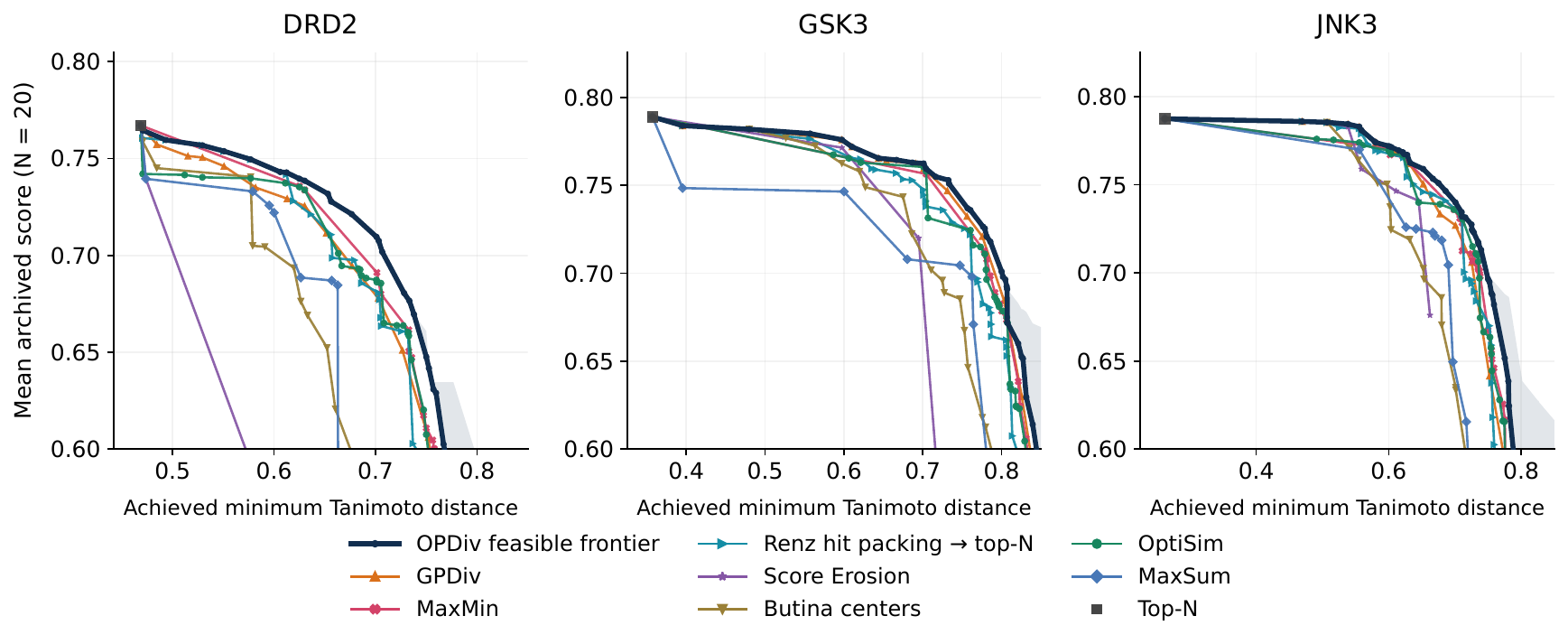}
\caption{Best observed score--diversity trade-offs for 20 compounds, using the first repeat of each target. Moving right requires greater minimum pairwise Tanimoto distance; moving up gives a higher mean score. Colored curves show the selection methods across their parameter sweeps. The dark OPDiv reference combines allowed selections from optimization and all selectors; shading shows the gap to the upper bound for the full pool at evaluated settings. Lines connect evaluated points as a guide to the eye. Panel axes differ; score axes start at 0.6 for readability. All ordinary top-$N$ points are shown.}
\label{fig:selector-frontiers}
\end{figure*}

We compare methods for selecting $N=20$ compounds without a minimum-score filter. Each method receives the same pool. Figure~\ref{fig:selector-frontiers} shows seven selection methods and ordinary top-$N$. Each curve shows the best score--distance trade-offs found across the tested settings and random seeds. Separation is the smallest pairwise Tanimoto distance in the selected set. We report a mean only for complete size-20 selections.

The full experiment covers 17 method families and 324,616 combinations of parameters and random seeds, with and without a minimum score of 0.5. We compute bounds for the full pools at 1,521 threshold evaluations, optimizing over the top 500, 1,000, and 2,000 candidates by score with a five-second solver limit for each subset. These limits exclude preprocessing and model construction. We choose the best settings on these same pools; computational budgets vary across methods. Appendix~\ref{app:protocol} records the displayed methods' settings and numerical conventions.

\subsection{How much score can better selection recover?}
\label{app:selector-comparison}
Several selectors approach the best attainable mean over parts of the distance range, particularly for GSK3 (Figure~\ref{fig:selector-frontiers}). Little score can be recovered there under the chosen pairwise rule. Elsewhere, larger gaps show that a better combination is available in the same pool. The shaded gap above the OPDiv reference shows how much further improvement may be possible.

To compare methods at the same minimum distance, we use each method's highest mean among its saved size-20 selections that meet the requirement. We calculate score gains separately for each run, then average them for each target.

At a minimum Tanimoto distance of 0.70, optimized selection improves the mean score over GPDiv by 0.013 for DRD2, 0.004 for GSK3, and 0.018 for JNK3, averaged over four repeats. Gains over Butina centers are 0.124, 0.049, and 0.106. These score gains are certified within $10^{-6}$. Ordinary top-$N$ does not meet this distance requirement in any of the 12 pools.

These comparisons use a minimum-distance requirement, while MaxSum and Score Erosion optimize other criteria. The Butina baseline selects cluster centers; CPDiv selects the best-scoring members under fixed cluster limits. Some score loss can arise when a tested setting produces more separation than required or the parameter sweep misses a better setting.

\subsection{Scores within selected sets}
\label{sec:diversity-spectra}
\begin{figure*}[t]
\centering
\includegraphics[width=\textwidth]{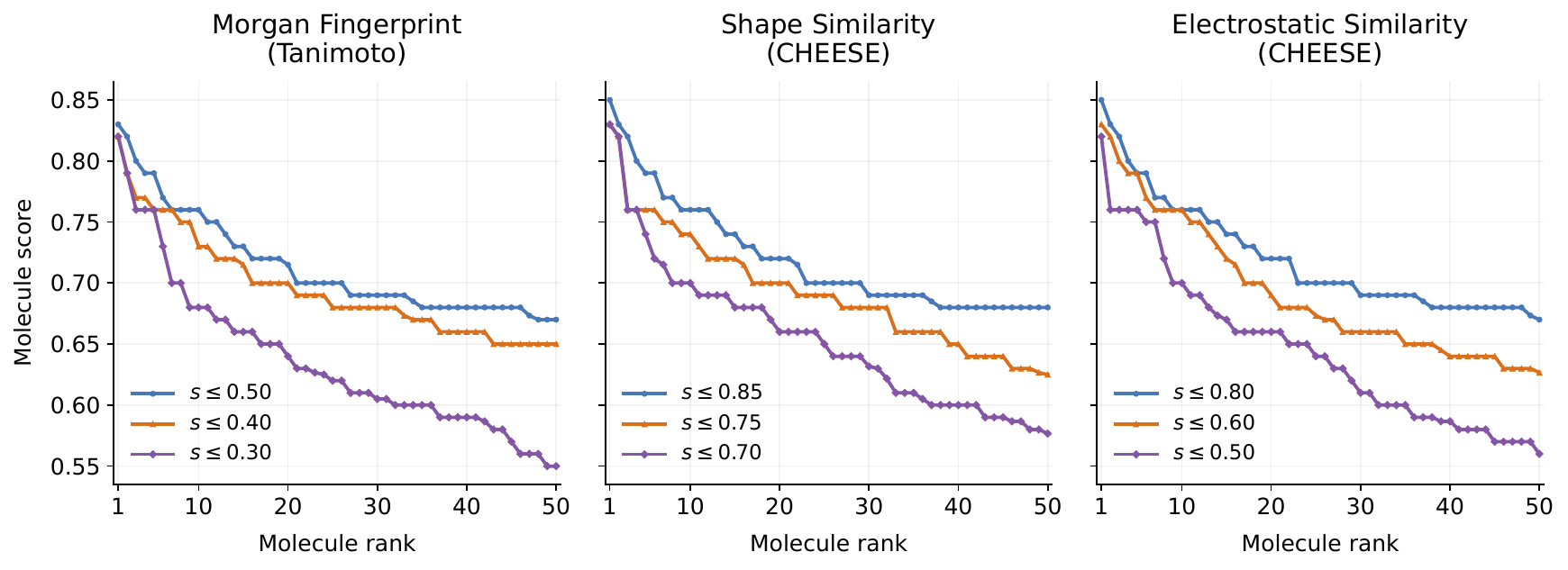}
\caption{Scores within separately optimized 50-compound sets from DRD2 repeat 1, using Morgan Tanimoto, CHEESE shape cosine, or CHEESE electrostatic cosine similarity. Every dot is a selected molecule, ordered by its score. Lower similarity limits require greater separation and can bring weaker-scoring compounds into the set. All nine mean scores are certified within $10^{-6}$. Thresholds and colors describe settings within each panel. Lines guide the eye.}
\label{fig:diversity-spectra}
\end{figure*}

Figure~\ref{fig:diversity-spectra} shows score profiles for 50-compound selections from all 10K eligible molecules in DRD2 repeat 1. The panels use Morgan Tanimoto similarity and cosine similarities from the original CHEESE \texttt{shapesim} and \texttt{espsim} SMILES encoders \citep{lzicar_cheese_2024}. These encoders map SMILES to learned representations of shape and electrostatics. The original DRD2 score is the selection objective in every panel.

Each curve is a separately optimized set under one of three similarity limits. We chose the limits after inspecting this pool to illustrate increasing separation within each representation. All nine means are certified within $10^{-6}$ over the full pool, using the top 500 candidates by score for optimization. Encoder and validation details are in Appendix~\ref{app:representations}.

As the limits become stricter, optimal means fall from 0.714 to 0.636 for Tanimoto, 0.720 to 0.654 for shape, and 0.720 to 0.645 for electrostatics. The lowest selected scores fall from 0.670 to 0.550, 0.680 to 0.577, and 0.670 to 0.560, respectively. Thus the profiles reveal weaker members needed to fill the more separated sets. Under matched budgets and selection rules, high scores sustained deeper into the profile would indicate a stronger supply of promising candidates.

\section{Discussion and applications}
\subsection{Matching the selection to the experimental decision}
OPDiv measures the quality of a shortlist for experimental follow-up. For a virtual-screening shortlist, $N$ can reflect the number of compounds that can be purchased or tested, the score can reflect the project's prioritization model, and the pairwise rule can limit redundant choices. A chemist can inspect the selected compounds to see which series enter, which high-scoring analogs are excluded, and what score is sacrificed for broader selection. Eligibility rules should remove unacceptable candidates before optimization.

The appropriate diversity rule depends on the next experiment. A one-per-series policy is useful when the intended decision is to advance different series; a fixed allowance of several analogs may better match a within-series follow-up. CPDiv makes that allowance explicit and has an exact, inexpensive selector for disjoint groups. Among sets that meet the diversity requirement, OPDiv chooses by mean score. Where extra separation is itself valuable, the full trade-off curve or another objective is needed.

The current formulation assumes equal follow-up costs. Unequal prices, shared synthetic routes, and batch chemistry require additional constraints or a different budget objective; SPARROW incorporates synthesis into compound selection \citep{fromer2025}. The meaning of the result depends on the score. For calibrated individual success probabilities, the sum gives the expected number of successes, even when outcomes are correlated. Docking and other surrogate scores provide a basis for prioritizing compounds.

\subsection{Evaluating and guiding molecular search}
A generator may return many close analogs yet still supply an excellent diverse shortlist. OPDiv gives that shortlist full credit, while the acquisition budget accounts for the effort spent obtaining it. Conversely, a widely dispersed output need not supply enough high-scoring compounds. Reporting best-set quality at fixed acquisition budgets, alongside score--diversity curves, would separate the ability to discover useful candidates from the ability to assemble them into a final selection.

At fixed $N$, OPDiv can plateau once a pool contains an excellent set; additional chemical-space coverage may then leave its value unchanged. Larger follow-up sizes ask a different question. Their optimal means cannot increase, because deleting a lowest-scoring member preserves pairwise compatibility, although the optimal sets need not be nested. Coverage measures describe the chemical space explored, while diverse-hit counts measure the number of separated candidates above a score cutoff.

A possible extension is to use improvement in best-set quality to guide candidate acquisition or reinforcement learning. After a feasible size-$N$ set exists, such a reward would credit a new candidate when it improves the best available set. This would require handling sparse rewards, repeated optimization costs, and score changes caused by solver approximations.

\subsection{Library selection and design}
The same decision can be posed for selecting DNA-encoded library hits for follow-up or choosing scored products from a make-on-demand library: which allowed set best uses the available experimental places? Chemical groups, scaffold novelty relative to a fixed reference collection, or learned similarity measures can express different selection priorities.

Designing an entire combinatorial library is a broader problem. Choosing a building block or reaction can add many products simultaneously. Shared reagents, reaction feasibility, synthetic cost, and coverage requirements would need explicit treatment. Extending OPDiv from selecting scored products to designing libraries would require linking product scores to the choice of building blocks and reactions.

\subsection{Robustness and limitations}
\label{sec:metric-gameability}
After deduplication, exact OPDiv values are independent of input order and duplicate records. Adding candidates preserves or improves the optimum when existing scores and pairwise relationships stay fixed. A generator can still exploit errors in the score, weaknesses in the representation, or a similarity boundary. Score--diversity curves, complementary representations, and inspection of selected structures help reveal these sensitivities.

The score scale also matters. Multiplying all scores by a positive number or adding the same constant to every score preserves optimal membership, but a nonlinear transformation can change which combinations are preferred even when individual rankings are unchanged. Comparisons therefore require scores with a common meaning and scale. Likewise, fingerprints, embedding models, clustering policies, and thresholds remain modeling choices.

Computation is another limitation. General pair-constrained selection is difficult, and constructing all pairwise comparisons requires quadratic work. The reported solver limits exclude representations, graph construction, and other preprocessing. The next tests should compare independently generated pools at matched acquisition budgets, challenge the criterion with adaptive optimization, and measure whether the selected sets improve experimental outcomes.

\section{Conclusion}
OPDiv asks which set of a given size achieves the highest scores while meeting an explicit diversity requirement. The same answer supplies both a compound shortlist and an evaluation of the pool that produced it. GPDiv provides a fast greedy reference; CPDiv gives exact selection when fixed, disjoint groups and upper limits express the desired allocation. Score--diversity curves show the price of stronger separation, while selected-set profiles expose what the mean hides. Together with feasibility checks and optimization bounds, these views distinguish the cost of requiring diversity from score unnecessarily lost during selection.

\section*{Software and data availability}
Source code and package links are provided on the first page. The molecules and scores come from the public diverse-hit benchmark \citep{renz2024}. We plan to release the reproduction code, selected molecules, parameter settings, checkpoint hashes, embeddings, and solver bounds separately. Appendix~\ref{app:baselines} records supporting definitions and protocols.

\section*{Acknowledgements}
The author thanks all colleagues, especially Radoslav Kriv\'{a}k, for valuable feedback and advice on the positioning of this paper.

\bibliographystyle{icml2025}
\bibliography{references}

\bigskip
\input{appendices}
\end{document}

%% file: appendices.tex
\appendix
\section{Definitions and reproducibility details}
\label{app:baselines}
\subsection{Related measures and selection rules}
\label{sec:circles}
\textbf{\#Circles} is the maximum number of mutually separated molecules at a chosen distance threshold \citep{xie2023}. Under the same boundary convention, a size-$N$ OPDiv selection exists exactly when this maximum is at least $N$. A heuristic packing gives only a lower bound.

\phantomsection\label{sec:sediv}
\textbf{Sphere-exclusion diversity} is the fraction of input molecules retained by sequentially accepting a molecule and excluding its neighbors \citep{thomas2021}. The result depends on input order, tie handling, deduplication, and sampling. \textbf{Internal diversity} is average pairwise dissimilarity \citep{benhenda2017}; including diagonal terms gives $m^{-2}\sum_{i,j}(1-s(i,j))$ for $m>0$ molecules. Close pairs can occur even when average dissimilarity is high. \textbf{Cluster counts} count nonempty groups under the chosen representation, algorithm, and threshold \citep{butina1999}.

\phantomsection\label{sec:diverse-hit-count}
\textbf{Diverse-hit counts} first apply a score cutoff, then count separated qualifying molecules \citep{renz2024}. A maximum count of at least $N$ means that every member of some allowed size-$N$ set meets that cutoff. An OPDiv mean above the cutoff can include members below it. A heuristic count provides a lower bound on the maximum.

\phantomsection\label{sec:dissimilarity-selectors}
\textbf{MaxMin} repeatedly adds the molecule farthest from its nearest selected neighbor; \textbf{MaxSum} instead maximizes the sum of distances to selected molecules \citep{ashton2002,borodin2012,gillet2011}. Both are greedy rules based on distances. \textbf{OptiSim} starts from a seed, excludes nearby candidates, samples up to $k$ remaining candidates uniformly without replacement, and chooses the sampled candidate farthest from its nearest selected neighbor \citep{clark1997}. With $k=1$, it gives random-order sphere exclusion. These selectors stop at $N$ or exhaustion, keeping any separation threshold fixed. GPDiv, ordinary top-$N$, and CPDiv are defined in the main text.

\subsection{Selector settings and matched comparisons}
\label{app:protocol}
MaxMin, MaxSum, and OptiSim are each applied to the top $M$ eligible molecules for $M=20$, 40, 100, 200, 500, 1,000, and the full pool, clipping oversized prefixes and removing duplicate settings. We apply this score filter before running each selector. MaxMin and MaxSum use 30 random starts and a highest-score start. OptiSim uses 30 seeds, $k\in\{1,10,100\}$, and exclusion distances 0 and 0.30--0.90 in steps of 0.05. Candidates are ordered by descending score, then canonical pool order; sampled ties follow this order.

GPDiv uses strict separation thresholds 0.30--0.90 in steps of 0.025. RDKit Butina uses the same numerical grid in its native distance convention, with and without dynamic reordering; the centers baseline keeps the 20 highest-scoring centers when available. Renz-style hit packing applies MaxMin separated-hit selection before taking the top 20 scores, using score cutoffs 0.5, 0.6, and 0.7, the same 25 separation thresholds, and 30 seeds. Score Erosion uses difference and product updates from the reference KNIME implementation, with 41 erosion factors from 0 to 1 \citep{meinl2010,meinl2011}.

OptiSim requires $d(i,j)>\delta+10^{-12}$. At achieved-distance comparisons, OPDiv instead allows $d(i,j)\geq\delta-10^{-12}$ so that the comparator's selected set remains feasible. Score-gain comparisons use saved selections that meet the stated distance requirement. Subtracting a qualifying comparator mean $H$ from a certified interval gives the gain interval $[L-H,U-H]$; endpoints are averaged over qualifying repeats within each target. A comparison is omitted when no tested setting produces a qualifying selection. Valid earlier certificates are reused only on identical indexed pools under matching scores and selection rules.

\subsection{Shape and electrostatic score profiles}
\label{app:representations}
Each original CHEESE encoder, \texttt{shapesim} or \texttt{espsim}, is loaded separately with \texttt{SentenceTransformer} and receives the original SMILES \citep{lzicar_cheese_2024}. It produces normalized 256-dimensional embeddings; all inputs fit within the checkpoints' 512-token limit. Cosine similarity is computed between embeddings. The similarity limits are $\{0.50,0.40,0.30\}$ for Tanimoto, $\{0.85,0.75,0.70\}$ for shape cosine, and $\{0.80,0.60,0.50\}$ for electrostatic cosine. Allowed pairs satisfy $s(i,j)\leq\tau+10^{-12}$. Scores, identities, set sizes, and every selected pair are independently checked; checkpoint hashes, embeddings, selected molecules, and bounds are retained with the experiment artifacts.

\subsection{Integer optimization and full-pool bounds}
\label{app:computation}
\label{app:input-policy}
Let $E$ list all forbidden pairs, and let $x_i=1$ when molecule $i$ is selected. The integer formulation is
\begin{equation}
\begin{aligned}
 \max_{x\in\{0,1\}^m}\quad &\frac1N\sum_i q_i x_i\\
 \text{subject to}\quad &\sum_i x_i=N,\\
 &x_i+x_j\leq1\quad(\{i,j\}\in E).
\end{aligned}
\label{eq:ip}
\end{equation}
Molecules and forbidden pairs form a conflict graph; allowed sets are independent sets \citep{qin2012}. Combining conflict lists by union forbids a pair rejected by any criterion; intersection forbids only pairs rejected by every criterion. Either rule must be declared. Deduplication must also specify how repeated measurements are handled: retaining the highest of repeated noisy scores can inflate the score.

For a prefix $P$ of the $M\geq N$ highest-scoring molecules, retain all within-prefix conflicts. If molecules are omitted, let $c$ be their highest score and add $N$ placeholders of score $c$, compatible with everything. Any full-pool selection maps to this relaxed problem by replacing omitted members with placeholders, without lowering its score or introducing conflicts. Its optimum therefore bounds the full-pool optimum from above. Final selections contain only molecules from the original pool.

CP-SAT uses one worker and seed 20260914. Pair constraints within each prefix are compressed into clique constraints: at most one member of each mutually conflicting group may be chosen, with every conflict edge covered. This preserves the allowed sets. Solver limits exclude graph and model construction.

Real scores use integer coefficients $w_i=\operatorname{round}(Kq_i)$ with $K=3{,}000{,}000$; placeholders use $\lceil Kc\rceil$. Let $e_{[1]},\ldots,e_{[N]}$ be the $N$ largest errors $|q_i-w_i/K|$ among real prefix molecules, and let $\varepsilon_N=N^{-1}\sum_{r=1}^N e_{[r]}$. If $B_{\rm int}$ bounds the relaxed integer objective sum and $T_N$ is the full-pool ordinary top-$N$ mean, then
\begin{equation}
 U=\min\left\{T_N,\frac{B_{\rm int}}{KN}+\varepsilon_N\right\}
 \label{eq:rounded-bound}
\end{equation}
is a valid original-score upper bound. Upward-rounded placeholders need no additional correction. With no omitted molecules, the same formula applies without placeholders. The lower bound $L$ is the highest original-score mean of a validated, allowed size-$N$ selection from optimization or any comparator.

If the solver returns an unknown status, we retain the top-$N$ upper bound and known feasible selections. A result is certified within $10^{-6}$ when $U-L\leq10^{-6}$ in the original score units. Approximate neighbor retrieval that misses conflicts, or extra restrictions that remove allowed sets, require separate justification before their bounds can certify the intended problem.

\section{Weighted evaluation and random-equivalent acquisition}
\subsection{Connection to cluster-average enrichment}
\label{app:weighted-enrichment}
For a labeled reference library of $M$ compounds, let $y_i\in\{0,1\}$ indicate experimental activity. Fix a partition into chemotypes, let $J>0$ count groups containing actives, and let $a_j>0$ count the actives in such group $j$. For a selected size-$N$ set $S$, write $h_j(S)$ for its active count from group $j$. Cluster-average enrichment at fraction $N/M$ is \citep{mackey2009}
\begin{equation}
 \mathrm{EF}_{\mathrm{CA}}(S)
 =\frac{M}{NJ}\sum_{j=1}^{J}\frac{h_j(S)}{a_j}.
 \label{eq:cluster-average-enrichment}
\end{equation}
Each active contributes $1/a_{g(i)}$ to the sum and each inactive contributes zero, including those in groups with no actives. Thus the criterion is additive in selected compounds. With $q_i=y_i$ and $\omega_j=1/a_j$ for active groups (zero for groups with no actives), its best feasible value equals $(M/J)Q_N^{(\omega)}$. This is an algebraic connection to the established weighting scheme \citep{clark2008,mackey2009}, not a prospective selection rule using unknown outcomes. In a validation experiment, labels are withheld during selection and the resulting set is evaluated afterward.

For a uniformly sampled size-$N$ subset, $\mathbb E[h_j]=Na_j/M$, so $\mathbb E[\mathrm{EF}_{\mathrm{CA}}]=1$. A random selector subject to diversity constraints generally has a different distribution and need not have this expectation. Inverse frequencies of all molecules in structural groups are an alternative priority, not the same correction as inverse active counts.

Equation~\ref{eq:weighted-opdiv} averages weighted utilities with the fixed denominator $N$; it does not divide by the sum of selected weights. This preserves linearity. Nonuniform weights make the origin of the original score consequential: adding a constant to every $q_i$ need not preserve the selected set because total selected weight can vary. Scores and weights must therefore express a declared utility convention. Existing full-pool bounds apply after replacing the objective coefficients and computing prefix orders, omitted-candidate bounds, and rounding errors from those weighted coefficients.

\subsection{Estimating the random matching budget}
\label{app:random-equivalent}
Fix the reference library, scores, selection rules, and target $q_\star$ before random sampling. For a uniform random permutation of the $M$ reference compounds, define the success event
\begin{equation}
\begin{aligned}
 \mathcal H_m(q_\star)=\bigl\{&\exists S\subseteq R_m:\ |S|=N,\\
 &S\text{ allowed},\quad N^{-1}\!\sum_{i\in S}q_i\geq q_\star\bigr\}.
\end{aligned}
\end{equation}
Then $T(q_\star)=\min\{m:\mathcal H_m(q_\star)\}$, with $T=\infty$ when no reference portfolio reaches the target. Infeasible prefixes are failures of this event, without assigning them an OPDiv score. If the target is reachable in the full reference library, every permutation reaches it by $M$, and the tail-sum identity gives
\begin{equation}
 \mathbb E[T(q_\star)]
 =\sum_{m=0}^{M-1}\Pr[T(q_\star)>m].
 \label{eq:matching-budget-tail}
\end{equation}

An estimator draws independent permutations and locates $T$ for each. Success is monotone in prefix size because every previously available portfolio remains available. After establishing reachability in the full library, binary search therefore needs only $O(\log M)$ prefix checks per permutation. Each check asks for a feasible selection under Equation~\ref{eq:ip} with the additional threshold $\sum_i q_ix_i\geq Nq_\star$; it need not determine the prefix optimum. The sample mean estimates the expected matching budget, and empirical quantiles estimate Equation~\ref{eq:matching-budget-quantile}. Report Monte Carlo uncertainty, for example by resampling independent permutation results, conditional on the fixed reference library and target.

A validated greedy portfolio reaching the target certifies success. An unconstrained top-$N$ mean below the target certifies failure. Only unresolved cases require a general solver, and precomputed conflict relationships can be reused. General pairwise-constrained selection remains computationally difficult \citep{qin2012}; logarithmically many checks do not imply polynomial total running time. Integer-score rounding must be handled through original-score validation and valid bounds as in Appendix~\ref{app:computation}. A timeout is unresolved, not evidence of failure. Certified failed and successful prefixes bound $T$; unresolved intervals should be retained rather than discarded or replaced by exact hitting times.

For CPDiv with fixed disjoint groups and upper capacities alone, each prefix optimum is obtained exactly by greedy selection. After one score sort, a check scans that order, skipping compounds outside the prefix or groups at capacity. This gives $O(M)$ work per check and $O(M\log M)$ work per permutation using binary search, without general integer optimization. The same procedure applies to fixed weighted coefficients. These estimates assume reference scores and the required group assignments or pairwise relationships are available; obtaining them adds scoring and preprocessing costs.

For an analytic check, take $N=1$ and suppose $H\geq1$ of the $M$ eligible reference compounds meet the score target. The absence of a qualifying compound from a uniform size-$m$ subset gives
\begin{equation}
\begin{aligned}
 \Pr[T\leq m]&=1-\frac{\binom{M-H}{m}}{\binom{M}{m}},\\
 \mathbb E[T]&=\frac{M+1}{H+1}.
\end{aligned}
 \label{eq:single-compound-matching}
\end{equation}
Here $0\leq m\leq M$, with a zero numerator when $m>M-H$. The expectation follows by symmetry of the $H+1$ gaps surrounding the qualifying compounds in a random permutation. If $H=0$, the target is unreachable.

\subsection{Interpretation and reporting}
The sampling factor is conditional on the candidate pool's target and the declared random-acquisition policy. A shared source library permits a comparison of acquisition methods; sampling from the candidate pool itself instead measures how many of its members are needed to recover its portfolio quality. Sampling is without replacement in the finite-library definition. Nonuniform sampling, repeated or ineligible proposals, and unequal evaluation costs require explicit accounting under the same policy for the method and its baseline.

If the candidate optimum is bounded by $L_A\leq Q_N(A;\tau)\leq U_A$, monotonicity in the target gives corresponding matching-budget bounds at $L_A$ and $U_A$, with a potentially unreachable upper target. Alternatively, use the mean of the validated candidate portfolio as an explicitly attained target. Any matching tolerance must be declared in the score units. For equal weights, positive affine transformations of all scores preserve the matching events when the target and any tolerance are transformed consistently.

Unlike conventional enrichment under uniform random selection, Equation~\ref{eq:opdiv-enrichment} has no guaranteed expectation of one when the candidate pool is itself randomly generated. It is a ratio of acquisition budgets conditional on attained quality, not a significance test or an estimate of biological enrichment. A factor of eight could, for example, represent 1,000 candidate acquisitions versus an estimated mean of 8,000 random acquisitions; this is an illustrative interpretation, not an experimental result. The weighted and random-equivalent extensions require separate empirical evaluation beyond the experiments reported here.